\documentclass{article}

\usepackage{microtype}
\usepackage{graphicx}
\usepackage{subcaption}
\usepackage{booktabs} 

\usepackage{hyperref}

\usepackage[preprint]{icml2026}

\usepackage{amsmath}
\usepackage{amssymb}
\usepackage{mathtools}
\usepackage{amsthm}
\usepackage{xurl}
\newcommand{\fullcov}{\ensuremath{\checkmark}}
\newcommand{\partialcov}{\ensuremath{\circ}}
\newcommand{\nocov}{\ensuremath{\times}}

\usepackage[most]{tcolorbox}
\usepackage{xcolor}

\definecolor{takeawayblue}{RGB}{49, 108, 168}
\definecolor{takeawaybg}{RGB}{239, 246, 252}

\newcounter{takeaway}

\newtcolorbox{takeawaybox}{
    enhanced,
    breakable,
    colback=takeawaybg,
    colframe=takeawayblue,
    boxrule=0pt,
    leftrule=0pt,
    rightrule=0pt,
    toprule=0.7pt,
    bottomrule=0.7pt,
    sharp corners,
    left=3pt,
    right=3pt,
    top=3pt,
    bottom=3pt,
    before skip=5pt,
    after skip=5pt,
    fontupper=\small
}

\newenvironment{takeaway}[1]
{
    \refstepcounter{takeaway}
    \begin{takeawaybox}
    \textcolor{takeawayblue}{
        \textbf{Takeaway \thetakeaway: #1}
    }\enspace
}
{
    \end{takeawaybox}
}

\usepackage[capitalize,noabbrev]{cleveref}

\theoremstyle{plain}

\theoremstyle{definition}

\theoremstyle{remark}

\usepackage[textsize=tiny]{todonotes}

\icmltitlerunning{World Model Interpretability: From Decoding Representations to Understanding Latent Worlds}

\begin{document}

\twocolumn[
  \icmltitle{World Model Interpretability: From Decoding Representations to Understanding Latent Worlds}



  \icmlsetsymbol{equal}{*}

  \begin{icmlauthorlist}
    \icmlauthor{Jize Wang}{sjtu,ntu}
    \icmlauthor{Yongcheng Jing}{ntu,equal}
    \icmlauthor{Xinyi Le}{sjtu}
    \icmlauthor{Cailian Chen}{sjtu,equal}
    \icmlauthor{Xinping Guan}{sjtu}
    \icmlauthor{Dacheng Tao}{ntu,equal}
  \end{icmlauthorlist}

  \icmlaffiliation{sjtu}{School of Automation and Intelligent Sensing, Shanghai Jiao Tong University}
  \icmlaffiliation{ntu}{College of Computing and Data Science, Nanyang Technological University}

  \icmlcorrespondingauthor{Yongcheng Jing}{yongcheng.jing@ntu.edu.sg}
  \icmlcorrespondingauthor{Cailian Chen}{cailianchen@sjtu.edu.cn}
  \icmlcorrespondingauthor{Dacheng Tao}{dacheng.tao@gmail.com}
    
  \icmlkeywords{Machine Learning, ICML}

  \vskip 0.3in
]

\newcommand{\yc}[1]{\textcolor{red}{#1}}
\newcommand{\ycj}[1]{\textcolor{red}{\bf [Comments: #1] }}

\newcommand{\revision}[1]{\textcolor{blue}{#1}}



\printAffiliationsAndNotice{}  

\begin{abstract}

Emerging Joint-Embedding Predictive Architectures (JEPAs) are moving beyond representation learning towards latent world models that support prediction, planning, and control. 
Yet their latent worlds naturally lack a human-interpretable interface. 
In this Perspective, we argue that as JEPA latents become internal states for prediction and action, interpretability must move beyond what they encode to how latent variables interact, evolve, and causally drive prediction.
To this end, we envision a four-stage research agenda, involving Ground, Structure, Verify, and Act, to guide this shift toward actionable latent-world understanding.
Within our framework, existing JEPA interpretability efforts provide early evidence for Ground, while systematic analysis of latent structure, causal verification, and intervention remains largely unexplored.
Overall, this Perspective aims to call for a broader latent-world interpretability agenda that shifts attention from decoding representations toward understanding, validating, and acting on latent worlds.
\footnote{A curated list of related work is available at \url{https://github.com/Jize-W/Awesome-World-Model-Interpretability}.}
\end{abstract}

\section{Introduction}
\label{sec:intro}

Learning models that can understand and predict the physical world is a long-standing goal of artificial intelligence~\citep{wmsurvey,dreamerv3,wm}. 
In particular, the emerging Joint-Embedding Predictive Architectures (JEPAs)~\citep{ijepa,vjepa,mcjepa,seqjepa,graphjepa} offer a promising approach by predicting in representation space rather than reconstructing observations at the pixel level. 
Recent notable developments such as V-JEPA 2~\citep{vjepa2,lewm} further extend this idea toward world modeling, where learned latent representations support the prediction of future states and, when conditioned on actions, can be used for planning and control. JEPA-based models have also expanded beyond general-purpose visual representation learning to embodied control and robotics~\citep{vla-jepa,psg-jepa}, 3D scene understanding~\citep{3d-jepa,3d-jepa-hu,ad-l-jepa}, medical analysis~\citep{mmjepa-ecg,us-jepa,jhu-vpt}, scientific discovery~\citep{jepa-dna,cell-jepa,hep-jepa}, and other domains~\citep{mts-jepa,6g-jepa}. As JEPAs take on broader roles, their latent spaces increasingly serve as internal states for representing and predicting how the world evolves.

This encouraging shift, on the other hand, also raises a largely overlooked question: \textit{what world has a JEPA learned inside its latent space?} 
This question is challenging, because unlike generative world models that produce human-observable predictions, JEPAs intrinsically predict in an abstract representation space without a naturally interpretable interface. 
When a prediction fails, it is therefore difficult to determine whether the failure arises from the represented state, learned dynamics, missing interactions, or errors accumulated over time. More broadly, predictive performance alone cannot establish whether a model has learned meaningful world structure or relies on shortcuts that succeed within its training distribution~\citep{howfar,whenlearn}.

\begin{figure*}[t]
    \centering
    \includegraphics[width=0.99\textwidth]{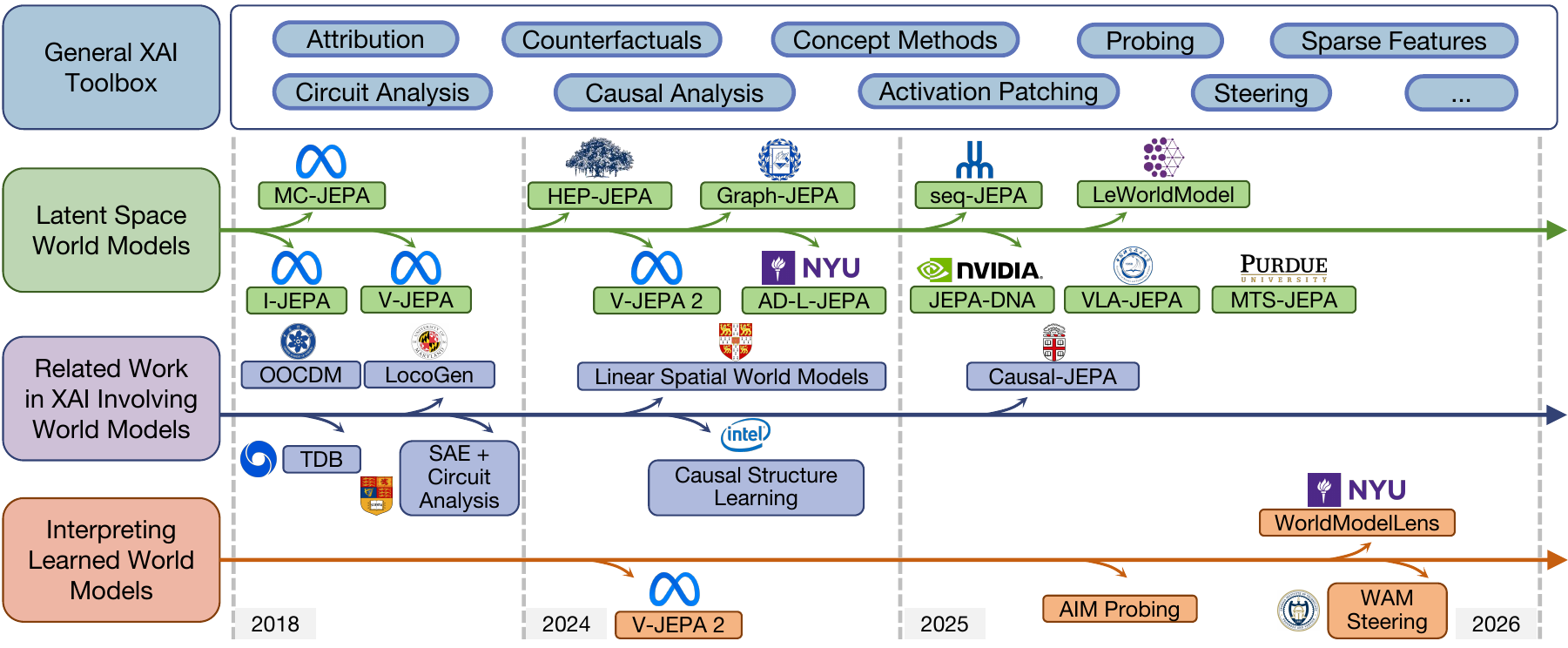}
    \caption{Evolution and positioning of interpretability for latent-space world models. Despite rapid advances in latent-space world modeling and a rich XAI toolbox, direct interpretation of learned world models remains sparse. Existing efforts largely focus on latent-content probing or isolated analysis and intervention techniques, leaving systematic understanding of latent worlds underexplored.}
    \label{fig:timeline}
\end{figure*}

Currently, few existing approaches to representation interpretability provide a natural starting point for addressing this problem. They can reveal what semantic or physical information is recoverable from learned representations, such as objects, motion, and actions~\citep{revisiting,linearspatial,sae+circuit}. However, decodability alone does not establish how this information participates in the model's internal world.  Recovering velocity, for example, does not show how it relates to other variables, shapes latent dynamics, or causally contributes to prediction. 
As such, interpreting a latent world model therefore requires moving beyond what its representations encode toward how they function together as a predictive system.

To instantiate this point, Figure~\ref{fig:timeline} situates the gap within the broader landscape of world modeling and interpretability. Although general XAI and mechanistic interpretability offer a rich toolbox~\citep{attribution,counterfactual,probe,concept,sae,circuit,causal,steering}, this toolbox has rarely been applied to learned world models as the primary object of interpretation. Existing XAI studies involving world representations largely pursue different objectives. 
Some introduce explicit world structure as a modeling mechanism: TDB~\citep{tdb} uses discrete bottlenecks to extract cognitive maps, OOCDM~\citep{oocdm} builds object-oriented causal dynamics models, and C-JEPA~\citep{causaljepa} introduces object-centric latent structure. 
Others use world models as an interpretability lens, asking whether Transformers or LLMs develop structured internal representations that resemble world models~\citep{causal-structure,sae+circuit,linearspatial}. In both cases, the learned world model itself is not the primary object of interpretation. Direct interpretability of learned world models remains scarce. Existing efforts primarily probe what information is encoded in latent representations~\citep{vjepa2,aim} or explore individual analysis and intervention techniques~\citep{worldmodellens,wam_steering}. Consequently, how latent variables are organized, evolve, and causally support prediction remains largely unexplored.

Our position is therefore: 
\emph{as JEPA latents become internal states for prediction and action, interpretability must move beyond what they encode to how latent variables interact, evolve, and causally drive prediction.}
Towards this shift, we envision a four-stage blueprint: \textbf{\textit{Ground, Structure, Verify, and Act}}. 
Specifically, Ground connects latent factors to variables in the external world. Structure studies how these factors relate and evolve through the model's learned dynamics. Verify tests whether the identified factors and structures are causally involved in prediction through controlled interventions. Finally, Act uses this understanding to diagnose failures and improve model behavior. Together, these stages shift the interpretability from identifying information that can be extracted from representations to understanding the internal mechanisms through which a latent world model represents, predicts, and acts upon the world.

In sum, our contribution is to establish a clear formulation of latent-world interpretability for predictive world models and provide a systematic blueprint for studying it. We organize existing efforts through this lens, identify the key gaps that remain, and outline how closing these gaps could enable more informative diagnosis, evaluation, and improvement of predictive world models. Finally, we discuss alternative views and identify priorities for advancing this emerging research agenda.

\section{Technical Setup}
\label{sec:technical_setup}

\subsection{JEPA as a Latent Predictive World Model}
\label{sec:jepa_world_model}

We consider a JEPA-based latent world model
\begin{equation}
    M = (E, P),
\end{equation}
where $E$ is an encoder and $P$ is a predictive dynamics model. Given an observation $x_t$ at time $t$, the encoder maps it to a latent state
\begin{equation}
    z_t = E(x_t),
\end{equation}
which represents the observed world internally. The predictor operates in this latent space to predict future states. In the action-conditioned setting,
\begin{equation}
    \hat{z}_{t+1} = P(z_{\leq t}, a_{\leq t}),
\end{equation}
where $a_{\leq t}$ denotes the action history up to time $t$. For action-free models, the action input is omitted.

Rather than predicting future observations directly in pixel space, the JEPA objective predicts their representations, allowing the model to focus on predictable aspects of the world. When used as a world model, $z_t$ and $P$ together define a latent predictive process that represents the current state and predicts its evolution.

We use \textbf{latent world} to refer to the internal variables, relations, and dynamics realized through this predictive process. The central object of interpretation is therefore not only an individual representation $z_t$, but the predictive system that transforms latent states. These transformations unfold over time and may also depend on actions.

To relate this latent world to the physical world, we describe the external world at time $t$ by a set of interpretable world variables
\begin{equation}
    \mathbf{c}_t =
    (c_t^1, c_t^2, \ldots, c_t^K),
\end{equation}
which may include object identity, position, velocity, geometry, contact, actions, events, and other properties.

\subsection{Interpretability under the Decoding Paradigm}
\label{sec:decoding_paradigm}

A common approach to representation interpretability asks what information about the world can be recovered from latent representations. Given a latent state $z_t$, a decoder $D$ seeks to recover the corresponding world variables:
\begin{equation}
    \hat{\mathbf{c}}_t = D(z_t).
\end{equation}
For a particular world variable $c^k$, with value $c_t^k$ at time $t$, one may learn a decoder
\begin{equation}
    \hat{c}_t^k = D_k(z_t)
\end{equation}
and evaluate how accurately $c_t^k$ can be recovered from the latent representation.

Linear and nonlinear probes test the decodability of particular variables, while concept-based methods, clustering, sparse representations, and related approaches identify correspondences between latent features and interpretable properties. Despite their methodological differences, these approaches share a common question: \textit{What information about the world can be decoded from the latent representation?}

We refer to this perspective as the \textbf{decoding paradigm}. It characterizes the semantic and physical content accessible from learned representations and provides a natural starting point for JEPA interpretability.

\subsection{The Gap: Decoding Is Not Understanding}
\label{sec:decoding_gap}

Decodability establishes that information about a world variable is accessible from a latent state, but not how it is organized or used by the predictive dynamics. This distinction matters for world models, where latent states are transformed over time to predict how the world evolves.

Consider a simple example in which velocity can be accurately decoded from $z_t$. This establishes that velocity-related information is accessible from the representation, but not how velocity relates to other variables such as object position or contact, or how these variables jointly evolve over time. More importantly, successful decoding does not establish that the predictor actually uses velocity-related information when computing $\hat{z}_{t+1}$~\citep{tdb}. The same prediction could instead be driven by correlated visual cues or other shortcuts.

We therefore distinguish \emph{representational accessibility} from \emph{world-model understanding}. The former concerns what information can be recovered from latent states; the latter concerns how that information is organized, evolves through learned dynamics, and causally contributes to prediction. For action-conditioned models, it also concerns how actions influence this evolution. The next section develops a four-stage blueprint for moving toward such understanding.

\begin{takeaway}{Decodability is not understanding.}
Recovering a world variable from a latent state shows that the information is accessible, but not how it is organized, how it evolves, or whether the predictor causally relies on it.
\end{takeaway}

\section{Blueprint: From Decoding to Understanding Latent Worlds}
\label{sec:blueprint}

To move from representational accessibility toward world-model understanding, we introduce a four-stage blueprint: \textbf{Ground, Structure, Verify, and Act}. The stages progressively ask what latent factors correspond to in the external world, how they relate and evolve, whether the model causally relies on them, and how verified understanding can be used to diagnose and improve model behavior.

Using the notation from Section~\ref{sec:technical_setup}, we denote sequences of latent states, world variables, and actions by
\begin{equation}
    \mathcal{Z} = \{z_t\}_{t=1}^{T}, \qquad
    \mathcal{C} = \{\mathbf{c}_t\}_{t=1}^{T}, \qquad
    \mathcal{A} = \{a_t\}_{t=1}^{T},
\end{equation}
where $\mathcal{A}$ is omitted for action-free models.

\begin{figure*}[t]
    \centering
    \includegraphics[width=0.99\textwidth]{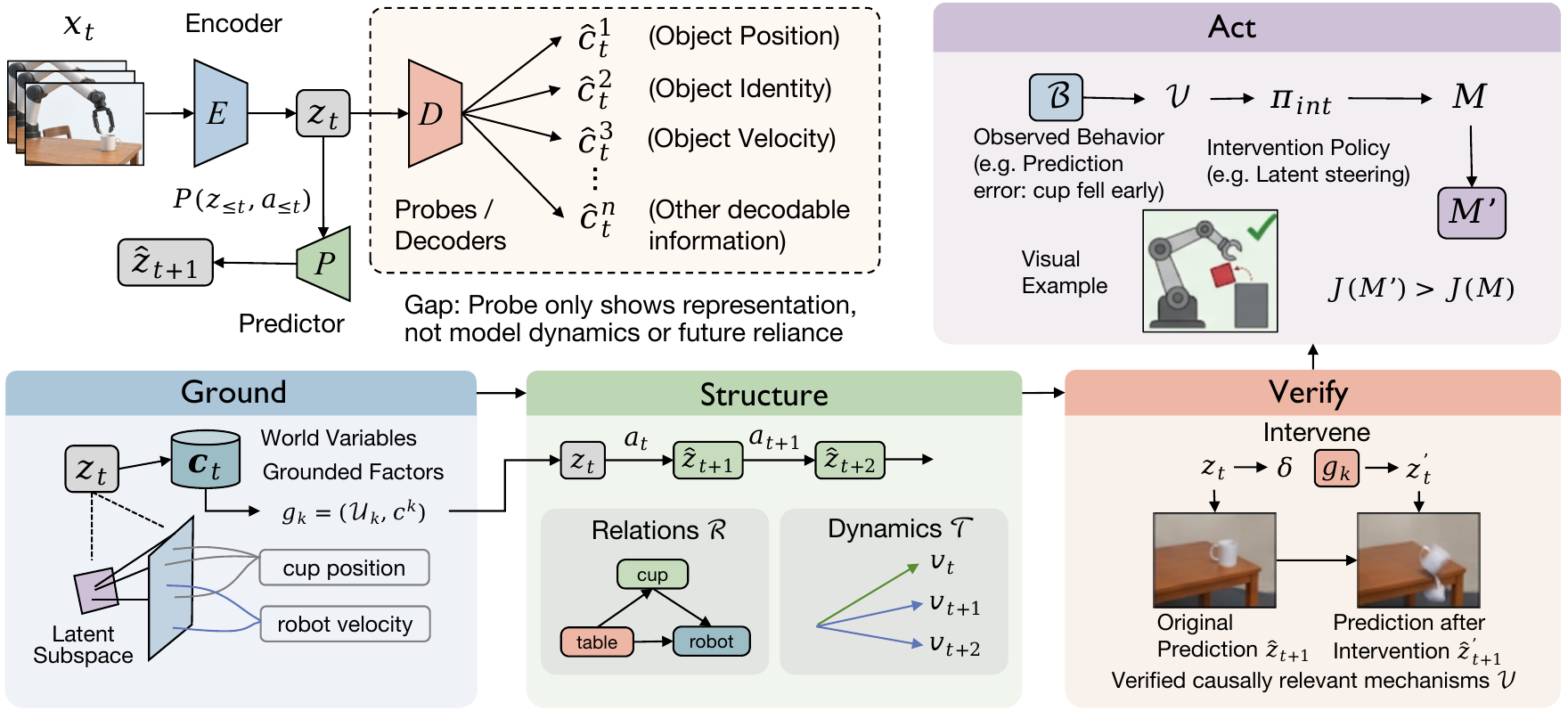}
    \caption{From decoding representations to understanding latent worlds. The proposed blueprint progresses through four stages: Ground connects latent factors to interpretable world variables; Structure recovers their relations and dynamics; Verify tests their causal roles through controlled interventions; and Act uses verified understanding to diagnose failures and improve model behavior.}
    \label{fig:framework}
\end{figure*}

\subsection{Ground: Connecting Latents to World Variables}
\label{sec:ground}

The first stage establishes correspondences between latent representations and interpretable world variables. Given latent states $\mathcal{Z}$ and world variables $\mathcal{C}$, we define
\begin{equation}
    \mathcal{G}
    \leftarrow
    \operatorname{Ground}(\mathcal{Z}, \mathcal{C}),
\end{equation}
where
\begin{equation}
    \mathcal{G} = \{g_k\}_{k=1}^{K},
    \qquad
    g_k = (\mathcal{U}_k, c^k),
    \qquad
    \mathcal{U}_k \subseteq \mathcal{Z}.
\end{equation}
Each $g_k$ associates a latent factor or subspace $\mathcal{U}_k$ with a world variable $c^k$:
\begin{equation}
    \mathcal{U}_k
    \quad \longleftrightarrow \quad
    c^k.
\end{equation}

Linear probes~\citep{probing}, concept probes~\citep{concept}, clustering~\citep{cluster}, sparse autoencoders (SAEs)~\citep{sae}, representation similarity, and related methods can be viewed as instances of Ground~\citep{interpretable_diffusion,dlmscope}. Despite their methodological differences, they seek correspondences between latent structure and interpretable properties of the external world.

\subsection{Structure: Recovering Relations and Dynamics}
\label{sec:structure}

Ground identifies latent factors associated with individual world variables, but does not reveal how they relate or evolve. Structure seeks to recover these relational and dynamical dependencies:
\begin{equation}
    (\mathcal{R}, \mathcal{T})
    =
    \operatorname{Structure}(\mathcal{G}, \mathcal{Z}, \mathcal{A}),
\end{equation}
where $\mathcal{R}$ captures relations among grounded factors and $\mathcal{T}$ captures their evolution. For action-conditioned models, $\mathcal{T}$ also captures how actions influence this evolution.

For example, after grounding factors associated with position, velocity, and contact, Structure asks how these factors interact and transition during processes such as motion or collision. Ground and Structure together define a hypothesis about the model's internal world:
\begin{equation}
    \mathcal{W}
    =
    (\mathcal{G}, \mathcal{R}, \mathcal{T}),
\end{equation}
where $\mathcal{W}$ captures its grounded variables, relations, and learned dynamics.  At this stage, however, $\mathcal{W}$ remains a hypothesis inferred from observed latent structure. Establishing that its components actually participate in the model's predictive computation requires causal verification.

\begin{figure*}[t]
    \centering
    \includegraphics[width=0.99\textwidth]{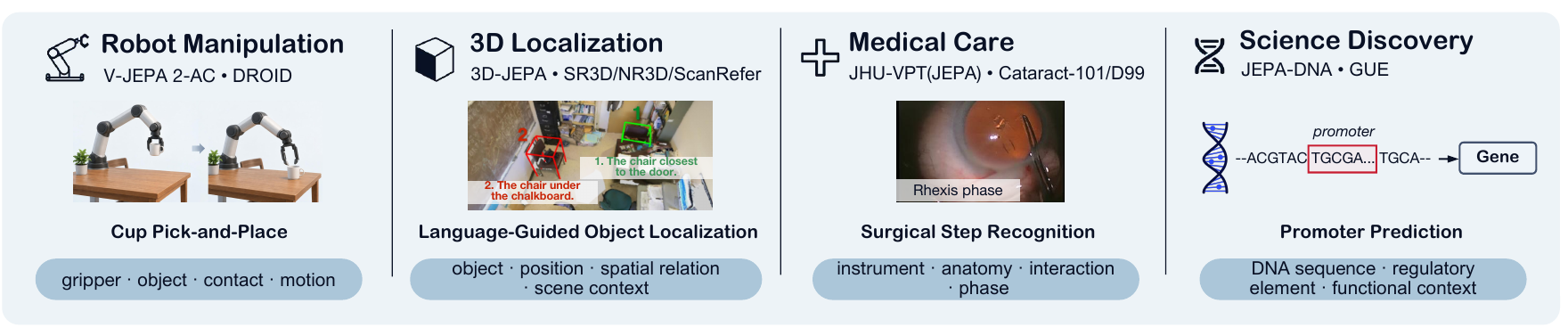}
    \caption{Cross-domain instantiations of latent worlds. Representative JEPA-based models across robot manipulation, 3D localization, medical care, and scientific discovery illustrate how relevant world variables, relations, and dynamics depend on the application domain. Latent-world interpretation therefore requires grounding representations in domain-specific concepts and interactions rather than assuming a universal set of interpretable features.}
    \label{fig:task}
\end{figure*}

\subsection{Verify: Testing Causal Roles}
\label{sec:verify}

Verify tests whether the hypothesized latent-world structure $\mathcal{W}$ causally participates in prediction. For a grounded factor $g_k \in \mathcal{G}$, consider a controlled intervention
\begin{equation}
    z_t'
    =
    \operatorname{Intervene}(z_t, g_k, \delta),
\end{equation}
where $\delta$ specifies a change to $g_k$. Its causal effect can be assessed by comparing
\begin{equation}
    \hat{z}_{t+1}
    =
    P(z_{\leq t}, a_{\leq t})
\end{equation}
with
\begin{equation}
    \hat{z}_{t+1}'
    =
    P(z_{\leq t}', a_{\leq t}).
\end{equation}
For action-free models, the action input is omitted.

More generally, given a set of controlled interventions or counterfactuals $\mathcal{I}$, we define
\begin{equation}
    \mathcal{V}
    =
    \operatorname{Verify}(M, \mathcal{W}, \mathcal{I}),
\end{equation}
where $\mathcal{V}$ contains the mechanisms in $\mathcal{W}$ whose causal roles are supported by intervention. Verification may target grounded factors in $\mathcal{G}$, relations in $\mathcal{R}$, or dynamics in $\mathcal{T}$.

For example, if a latent factor is grounded as velocity, changing it while holding other relevant factors fixed should systematically alter predicted motion. This distinguishes a factor that participates in prediction from one that is merely correlated with motion.

\subsection{Act: Using Interpretations to Change Model Behavior}
\label{sec:act}

The final stage uses verified understanding to diagnose and improve model behavior. Let $\mathcal{B}$ denote observed behaviors of interest, such as prediction errors, out-of-distribution failures, planning failures, or unsafe behavior. We define
\begin{equation}
    M'
    \leftarrow
    \operatorname{Act}(M, \mathcal{V}, \mathcal{B}).
\end{equation}

Given the verified mechanisms $\mathcal{V}$, Act may derive targeted interventions through latent steering, activation editing, retraining, fine-tuning, or related methods. Denoting such an intervention policy by $\pi_{\mathrm{int}}$, we have
\begin{equation}
    M' = \pi_{\mathrm{int}}(M).
\end{equation}

Its effect can be evaluated using a task-dependent criterion $J$:
\begin{equation}
    J(M') > J(M),
\end{equation}
where $J$ may measure prediction accuracy, out-of-distribution robustness, planning success, safety, or other relevant outcomes. Act thus closes the loop between interpretation and model behavior by turning verified knowledge of the latent world into a basis for diagnosis and targeted intervention.

Together, the four stages define a progression from representation-level interpretability toward latent-world understanding:
\begin{equation}
    (\mathcal{Z},\mathcal{C})
    \xrightarrow{\text{Ground}}
    \mathcal{G}
    \xrightarrow{\text{Structure}}
    \mathcal{W}
    \xrightarrow{\text{Verify}}
    \mathcal{V}
    \xrightarrow{\text{Act}}
    M'.
\end{equation}

\begin{takeaway}{Latent-world interpretability requires a progression from correspondence to causality and action.}
Ground identifies what latent factors correspond to; Structure reveals how they relate and evolve; Verify tests whether the predictor causally relies on them; and Act turns verified understanding into diagnosis and intervention.
\end{takeaway}

\section{Mapping Existing Work onto the Framework}
\label{sec:mapping}

\paragraph{From XAI to latent world interpretability.}
JEPA interpretability builds on a broader methodological foundation in explainable AI (XAI)~\citep{xai} and mechanistic interpretability~\citep{mi}. 
As described in Section~\ref{sec:intro} and Figure~\ref{fig:timeline}, existing methods range from feature attribution~\citep{attribution}, counterfactual explanations~\citep{counterfactual}, concept analysis~\citep{concept}, and probing to feature discovery~\citep{probe}, sparse autoencoders~\citep{sae}, circuit analysis~\citep{circuit}, and activation interventions~\citep{causal}. These methods provide many of the tools needed to instantiate our framework, particularly the progression from identifying internal features to causally testing and manipulating them.
However, latent predictive world models introduce a distinct object of interpretation: not only individual internal features, but also the relations and predictive dynamics that organize them into an evolving latent world. Applying existing interpretability tools in this setting therefore requires extending their focus from representations to the predictive system formed by those representations and their dynamics.

\begin{table*}[t]
\centering
\caption{
Representative work related to latent-world interpretability, mapped onto the proposed Ground, Structure, Verify, Act blueprint.
We focus on work that directly analyzes the representations, dynamics, or causal mechanisms of learned world models, with particular emphasis on JEPAs.
\fullcov\ explicit coverage;
\partialcov\ partial coverage;
\nocov\ not addressed.
}
\vspace{5pt}
\label{tab:mapping}
\setlength{\tabcolsep}{10pt}
\resizebox{0.9\linewidth}{!}{
\begin{tabular}{lllcccc}
\toprule
Role
& Model Scope
& Approach
& Ground
& Structure
& Verify
& Act \\
\midrule

Method
& JEPA
& V-JEPA 2 Probing
& \partialcov
& \partialcov
& \nocov
& \nocov \\

Method
& JEPA
& AIM Probing
& \partialcov
& \nocov
& \nocov
& \nocov \\

Infrastructure
& JEPA + Non-JEPA
& WorldModelLens
& \partialcov
& \nocov
& \partialcov
& \nocov \\

Transferable Method
& Non-JEPA
& WAM Steering
& \partialcov
& \partialcov
& \partialcov
& \fullcov \\

\bottomrule
\end{tabular}
}
\end{table*}

\paragraph{Latent worlds are domain-dependent.}
Figure~\ref{fig:task} illustrates how the relevant latent world depends on the application domain. In robot manipulation, relevant variables may describe objects, contacts, and motion~\citep{vjepa2,droid}; in 3D localization, objects, positions, and spatial relations become central~\citep{3d-jepa,sr3d,scanrefer}; while medical~\citep{jhu-vpt,cataract-101} and scientific settings~\citep{jepa-dna,gue} involve different domain-specific variables and interactions.

For example, in surgical video understanding~\citep{cataract-101}, an interpretable latent world may involve instruments, anatomical structures, their interactions, and the temporal progression of procedural phases. Decoding these concepts individually would not reveal whether the model captures the procedural structure connecting them or relies on clinically relevant factors when predicting subsequent events. The blueprint therefore provides a common interpretability framework while allowing the variables, relations, and dynamics under study to be defined by the domain.

\paragraph{Early evidence toward latent-world interpretability.}
Table~\ref{tab:mapping} summarizes the small body of work that is directly related to latent-world interpretability. Within JEPAs, probing in V-JEPA 2~\citep{vjepa2} shows that frozen representations contain information about appearance, motion, and action, while passive discrete probing~\citep{aim} examines correspondences between physical variations and unsupervised symbolic structure. These studies primarily provide evidence for grounding world-relevant information in JEPA latent spaces.

Emerging infrastructure and results from neighboring model classes begin to address later stages of the framework. WorldModelLens~\citep{worldmodellens} provides infrastructure for probing, attribution, sparse representation analysis, activation patching, and intervention replay across world-model architectures, including JEPAs. Work on World Action Models~\citep{wam_steering} provides a complementary precedent from a non-JEPA setting, where structured activation patterns are analyzed, dynamically modeled, and manipulated to affect downstream behavior. These developments suggest methodological paths from representational analysis toward causal intervention, although direct evidence in JEPA-based world models remains limited.

\paragraph{What remains missing.}
Existing work covers the proposed framework unevenly. Direct evidence in JEPAs is concentrated primarily at the Ground stage, with some emerging tools for causal analysis and intervention. Systematic study of how grounded variables are organized into relational and dynamical structure, whether that structure causally supports prediction, and how verified mechanisms can guide targeted intervention remains largely open.

\begin{takeaway}{Direct evidence for latent-world interpretability in JEPAs remains sparse.}
Existing JEPA interpretability is predominantly concentrated at Ground; advancing toward Structure, Verify, and Act represents an underexplored frontier for latent-world interpretability.
\end{takeaway}

\section{Implications: What Understanding Latent Worlds Enables}
\label{sec:implications}

In this section, we aim to shift the object of interpretation from representations to latent worlds, changing how world models can be analyzed and evaluated. 
If the internal variables, relations, and dynamics of a JEPA can be grounded and causally verified, they can help explain model successes and failures, assess when predictions should be trusted, and guide improvements in model behavior. 
To achieve this goal, we highlight four implications below.

\subsection{Diagnosing where prediction failures arise}

A prediction error alone reveals little about what went wrong inside a world model. Consider a robot pushing a cup near the edge of a table. If the model predicts that the cup will remain on the table but it instead falls, several internal failures could produce the same observed error. The latent state may fail to represent that the cup is near the edge (\emph{state representation}); the predictor may represent the state correctly but fail to capture the dynamics of losing support (\emph{dynamics}); it may fail to model the interaction among the robot, cup, and table (\emph{interaction}); or initially correct predictions may drift during multi-step prediction (\emph{rollout}).

Latent world understanding could make these failure modes distinguishable. Grounded factors can reveal whether the relevant state information is present, while recovered relations and transitions can expose where the predicted evolution departs from the expected dynamics. Causal analysis can further identify which internal mechanisms contribute to the failure. Thus, these capabilities could turn interpretability into a tool for localizing errors within the predictive process.

\subsection{Distinguishing learned world models from shortcuts}

Strong predictive performance does not necessarily imply that a model has captured the underlying structure of the world. Similar performance may arise from meaningful physical structure or from correlated cues such as background texture and camera motion. Latent world interpretability provides a way to distinguish these possibilities by examining whether verified internal mechanisms remain predictively relevant as such correlations change.

A model whose predictions consistently depend on stable world relevant factors across objects, scenes, and environments provides stronger evidence of having captured transferable world structure. Conversely, if prediction depends strongly on factors tied to incidental properties of the training distribution, the apparent world modeling capability may be more fragile than performance suggests. This distinction could provide a more informative basis for assessing what kind of world knowledge a JEPA has actually acquired.

\subsection{Explaining generalization and its limits}

Understanding latent worlds may also help explain why some predictive models generalize across environments while others fail under distribution shift. Rather than treating generalization only as an empirical property measured on downstream benchmarks, latent world analysis allows us to examine which parts of the internal model remain stable as the environment changes.

For example, latent factors that remain stable across changes in object appearance while varying systematically with geometry or dynamics may support transfer across visually different environments~\citep{disentangled}. Conversely, factors that change strongly with background, camera viewpoint, or other task-irrelevant properties may reveal likely sources of out-of-distribution failure~\citep{ad3,latent_dynamic}. Studying these invariances and dependencies could connect the internal organization of a latent world to the conditions under which its predictions remain reliable~\citep{recore}.

\begin{takeaway}{Similar predictive performance can hide fundamentally different internal worlds.}
Latent-world analysis can distinguish where failures arise, whether predictions rely on transferable world structure or shortcuts, and which internal mechanisms remain stable under distribution shift.
\end{takeaway}

\subsection{From post-hoc explanation to decision-bounded interpretation}

The consequences of latent world understanding become particularly important when model predictions are used for planning and control. In this setting, an inaccurate internal world can affect not only what the model predicts, but also which actions an agent chooses. Interpretability could provide an additional basis for assessing predictions before they are used for decision making, for example by identifying whether relevant state variables are represented reliably or whether the predicted dynamics are consistent with previously verified mechanisms.

This perspective also broadens the role of interpretability in safety. Rather than explaining failures only after an action has been taken, latent world analysis could help assess predictions before they are used for action. For agents that use JEPA predictions for planning and control, interpretability could reveal whether a prediction depends on internal variables and dynamics that have been reliably identified and verified, or on patterns known to be unstable under similar conditions. Such information could provide a signal for deciding when a model's prediction should be trusted.

\section{Alternative Views and Limitations}
\label{sec:alternative_views}

We consider several alternative views that challenge our position and help clarify its scope.

One possible view is that decoding may sometimes be sufficient. This is indeed the case for many interpretability questions. If the goal is to determine whether a JEPA representation contains information about object identity, motion, geometry, or another property, a well-designed probe may provide all the evidence that is needed. The need for deeper analysis arises when the question concerns the internal predictive world: how represented information is organized, how it participates in prediction, or whether the model causally relies on it. These questions require evidence beyond decodability. Our blueprint is intended for precisely this setting, where representation-level interpretation is a useful starting point but not the end of the analysis.

A deeper concern is whether human-defined variables provide the right ontology for interpreting a latent world. A JEPA need not decompose its internal computation into familiar concepts such as position, velocity, or contact~\citep{metaothello}. Its representations may instead be distributed, nonlinear, relational, or task-dependent, and some predictive factors may have no simple human-interpretable counterpart. Grounding should therefore not presuppose a one-to-one correspondence between human concepts and individual neurons or latent directions. Instead, such concepts can serve as reference points for discovering testable correspondences involving subspaces, nonlinear structures, or distributed representations. Cases in which the model's internal organization departs from human-defined variables may themselves reveal how it represents the world differently from our chosen description.

Another concern is that interventions in latent space may not always be valid. We agree that simply changing a latent factor is not sufficient to establish a causal relationship. A latent representation may encode multiple properties together, so changing a factor associated with velocity, for example, could unintentionally change other properties as well. The modified representation may also no longer correspond to a plausible state of the world. In either case, a change in prediction would be difficult to attribute to the intended intervention. An important challenge for Verify is therefore to develop interventions that change the target factor while preserving other relevant information and keeping the resulting latent state plausible. Determining how to construct and validate such interventions is an important direction for future research.

More broadly, interpretability may not be necessary for building reliable world models. Robust prediction and control can also be pursued through behavioral evaluation, uncertainty estimation, robustness training, and out-of-distribution testing without examining internal representations. Latent world interpretability should not replace these approaches. Its value is complementary: behavioral evaluation can reveal whether and under what conditions a model fails, while internal analysis may help identify the mechanisms behind those failures and distinguish models that achieve similar performance through different internal strategies. This additional evidence may be particularly useful when predictive models are used for planning and control, where prediction failures can directly affect subsequent decisions and actions.

Taken together, these considerations highlight important challenges of latent world interpretability. They also clarify where the proposed blueprint applies and where its assumptions and methods require further validation.

\begin{takeaway}{Going beyond decoding does not mean abandoning decoding or behavioral evaluation.}
Decoding remains sufficient for representation-level questions, while latent-world understanding is needed for claims about predictive mechanisms. Such claims require valid interventions and should complement, rather than replace, behavioral evaluation.
\end{takeaway}

\section{Call to Action}
\label{sec:call_to_action}

In aggregate, the aforementioned discussions indicate that realizing latent-world interpretability inevitably requires redirecting research efforts beyond isolated applications of existing interpretability tools toward a systematic research program centered on the variables, structures, dynamics, and causal mechanisms of learned worlds. To help shape future research, we highlight four priorities toward this goal.

\noindent\textbf{Build controlled testbeds and evaluation criteria.}
A first priority is to develop settings in which latent world interpretations can be tested against known properties of the world. Environments with known objects, physical variables, interactions, and dynamics provide a basis for evaluating whether recovered latent structure reflects the underlying system. Evaluation should go beyond decoding accuracy to test whether interpretations remain stable across contexts, whether recovered relations and dynamics are consistent with observed behavior, and whether causal claims hold under controlled changes. Such testbeds will not capture the full complexity of real-world prediction, but they can provide a foundation for evaluating latent world interpretations systematically.

\noindent\textbf{Study how latent worlds emerge and differ across models.}
Latent world interpretability should examine how internal predictive structure develops through learning. The organization learned by a JEPA may depend on its architecture, training objective, data, model scale, and stage of training~\citep{generalagent}. Comparing models along these dimensions could reveal when world-relevant factors and dynamics begin to emerge, whether similar structures arise across models, and which design choices lead to more stable or reusable internal structure. This would extend interpretability to studying how latent worlds form and evolve.

\noindent\textbf{Connect interpretation to model behavior and design.}
Interpretations should be tested against meaningful differences in prediction, generalization, failure modes, and downstream planning or control. Such connections can also inform model design. If particular internal structures are consistently associated with more reliable prediction or better generalization, they may provide useful evidence for comparing architectures, training objectives, or other design choices. 
Interpretability could thereby serve not only to analyze trained models, but also to understand which design choices lead to better predictive world models.

\noindent\textbf{Scale latent world analysis to realistic settings.}
Methods developed in controlled environments must eventually be tested on natural video, long-horizon prediction, and embodied agents operating in complex environments. Doing so will require tools for extracting and comparing internal states, tracing predictive computation over time, performing controlled interventions, and analyzing rollouts across model families. 
Shared infrastructure and evaluation protocols could make these analyses more reproducible and make it easier to compare findings across models and studies.



\section{Conclusion}

This Perspective aims to establish latent-world understanding as a central goal for interpreting predictive world models. We note that decodability is not understanding: accessible information does not reveal how it is organized, evolves, or causally supports prediction. Building on this observation, we envision the Ground–Structure–Verify–Act blueprint as a path toward this goal, and as a framework for shaping future research on latent-world interpretability. We call for future interpretability efforts to move beyond isolated representation-level analyses toward systematically understanding, validating, and acting on the internal worlds learned by predictive models.



\bibliography{example_paper}

@article{xai,
      title={Explainable artificial intelligence (XAI): from inherent explainability to large language models}, 
      author={Fuseini Mumuni and Alhassan Mumuni},
      journal={Array},
      year={2026},
}

@article{mi,
  title={Locate, steer, and improve: A practical survey of actionable mechanistic interpretability in large language models},
  author={Zhang, Hengyuan and Zhang, Zhihao and Wang, Mingyang and Su, Zunhai and Wang, Yiwei and Wang, Qianli and Yuan, Shuzhou and Nie, Ercong and Duan, Xufeng and Xue, Qibo and others},
  journal={Computer Science Review},
  volume={62},
  pages={101011},
  year={2026},
}

@article{vjepa2,
  title={V-jepa 2: Self-supervised video models enable understanding, prediction and planning},
  author={Assran, Mido and Bardes, Adrien and Fan, David and Garrido, Quentin and Howes, Russell and Muckley, Matthew and Rizvi, Ammar and Roberts, Claire and Sinha, Koustuv and Zholus, Artem and others},
  journal={arXiv:2506.09985},
  year={2025}
}

@article{aim,
  title={Probing the Latent World: Emergent Discrete Symbols and Physical Structure in Latent Representations},
  author={Ming, Liu Hung},
  journal={arXiv:2603.20327},
  year={2026}
}

@article{worldmodellens,
  title={One Lens, Many Worlds: A Capability-Typed Interface for World-Model Interpretability},
  author={Challagundla, Bhavith Chandra and Pandey, Sanskar and Thakkar, Param and Mallagundla, Rishikesh and Gogireddy, Yugandhar Reddy and Lu, Wenhao and Choudhury, Hindol Roy and Challagundla, Shravani and Nasr, Mohamed Deraz and Deshpande, Spursh},
  journal={arXiv:2606.09936},
  year={2026}
}

@article{wam_steering,
  title={Steering Robustness into World Action Models via Mechanistic Interpretability and Optimal Control},
  author={Hong, Jihoon and Skifstad, Julian and Dai, Qiyue and Chan, Alice and Chou, Glen},
  journal={arXiv:2607.14943},
  year={2026}
}

@article{wmsurvey,
	year = {2026},
	author = {Jiahua Dong and Qi Lyu and Baichen Liu and Xudong Wang and Wenqi Liang and Duzhen Zhang and Jiahang Tu and Hongliu Li and Hanbin Zhao and Henghui Ding and Yulun Zhang and Zhi Han and Nicu Sebe and Fahad Shahbaz Khan and Salman Khan and Mubarak Shan and Philip Torr and Ming-Hsuan Yang and Dacheng Tao},
	title = {Learning to Model the World: A Survey of World Models in Artificial Intelligence},
	journal = {Preprints}
}

@article{wm,
author = {Ha, David and Schmidhuber, Jürgen},
  
  title = {World Models},
  
  journal = {arXiv:1803.10122},
  year = {2018},
  
}

@article{dreamerv3,
  title={Mastering diverse control tasks through world models},
  author={Hafner, Danijar and Pasukonis, Jurgis and Ba, Jimmy and Lillicrap, Timothy},
  journal={Nature},
  volume={640},
  number={8059},
  pages={647--653},
  year={2025},
}

@inproceedings{ijepa,
  title={Self-supervised learning from images with a joint-embedding predictive architecture},
  author={Assran, Mahmoud and Duval, Quentin and Misra, Ishan and Bojanowski, Piotr and Vincent, Pascal and Rabbat, Michael and LeCun, Yann and Ballas, Nicolas},
  booktitle={CVPR},
  pages={15619--15629},
  year={2023},
}

@article{vjepa,
  title={Revisiting feature prediction for learning visual representations from video},
  author={Bardes, Adrien and Garrido, Quentin and Ponce, Jean and Chen, Xinlei and Rabbat, Michael and LeCun, Yann and Assran, Mahmoud and Ballas, Nicolas},
  journal={TMLR},
  year={2024}
}

@article{mcjepa,
  title={Mc-jepa: A joint-embedding predictive architecture for self-supervised learning of motion and content features},
  author={Bardes, Adrien and Ponce, Jean and LeCun, Yann},
  journal={arXiv:2307.12698},
  year={2023}
}

@inproceedings{seqjepa,
  title={seq-JEPA: Autoregressive predictive learning of invariant-equivariant world models},
  author={Ghaemi, Hafez and Muller, Eilif B and Bakhtiari, Shahab},
  booktitle={NeurIPS},
  volume={38},
  pages={32943--32973},
  year={2026}
}

@article{lewm,
  title={Leworldmodel: Stable end-to-end joint-embedding predictive architecture from pixels},
  author={Maes, Lucas and Lidec, Quentin Le and Scieur, Damien and LeCun, Yann and Balestriero, Randall},
  journal={arXiv:2603.19312},
  year={2026}
}

@inproceedings{howfar,
  title={How far is video generation from world model: A physical law perspective},
  author={Kang, Bingyi and Yue, Yang and Lu, Rui and Lin, Zhijie and Zhao, Yang and Wang, Kaixin and Huang, Gao and Feng, Jiashi},
  booktitle={ICML},
  year={2025}
}

@inproceedings{
whenlearn,
title={When do neural networks learn world models?},
author={Tianren Zhang and Guanyu Chen and Feng Chen},
booktitle={ICML},
year={2025},
}

@inproceedings{
revisiting,
title={Revisiting the Othello World Model Hypothesis},
author={Yifei Yuan and Anders S{\o}gaard},
booktitle={ICLR Workshop on World Models: Understanding, Modelling and Scaling},
year={2025},
}

@article{linearspatial,
  title={Linear spatial world models emerge in large language models},
  author={Tehenan, Matthieu and Moya, Christian Bolivar and Long, Tenghai and Lin, Guang},
  journal={arXiv:2506.02996},
  year={2025}
}

@inproceedings{interpretable_diffusion,
  title={Interpretable diffusion via information decomposition},
  author={Kong, Xianghao and Liu, Ollie and Li, Han and Yogatama, Dani and Ver Steeg, Greg},
  booktitle={ICLR},
  pages={55279--55292},
  year={2024}
}

@inproceedings{
dlmscope,
title={{DLM}-Scope: Mechanistic Interpretability of Diffusion Language Models via Sparse Autoencoders},
author={Xu Wang and Bingqing Jiang and Yu Wan and Baosong Yang and Lingpeng Kong and Difan Zou},
booktitle={ICML},
year={2026},
}

@inproceedings{
causaljepa,
title={Causal-{JEPA}: Learning World Models through Object-Level Latent Masking},
author={Heejeong Nam and Quentin Le Lidec and Lucas Maes and Yann LeCun and Randall Balestriero},
booktitle={ICML},
year={2026},
}

@inproceedings{disentangled,
  title={Disentangled world models: Learning to transfer semantic knowledge from distracting videos for reinforcement learning},
  author={Wang, Qi and Zhang, Zhipeng and Xie, Baao and Jin, Xin and Wang, Yunbo and Wang, Shiyu and Zheng, Liaomo and Yang, Xiaokang and Zeng, Wenjun},
  booktitle={ICCV},
  pages={2599--2608},
  year={2025},
}

@InProceedings{ad3,
  title = 	 {{AD}3: Implicit Action is the Key for World Models to Distinguish the Diverse Visual Distractors},
  author =       {Wang, Yucen and Wan, Shenghua and Gan, Le and Feng, Shuai and Zhan, De-Chuan},
  booktitle = 	 {ICML},
  pages = 	 {51546--51568},
  year = 	 {2024},
  volume = 	 {235},
  
}

@inproceedings{
latent_dynamic,
title={Learning Latent Dynamic Robust Representations for World Models},
author={Ruixiang Sun and Hongyu Zang and Xin Li and Riashat Islam},
booktitle={ICML},
year={2024},
}

@inproceedings{recore,
  title={Recore: Regularized contrastive representation learning of world model},
  author={Poudel, Rudra PK and Pandya, Harit and Liwicki, Stephan and Cipolla, Roberto},
  booktitle={CVPR},
  pages={22904--22913},
  year={2024},
}

@inproceedings{
metaothello,
title={MetaOthello: A Controlled Study of Multiple World Models in Transformers},
author={Aviral Chawla and Galen Hall and Juniper L Lovato},
booktitle={ICML},
year={2026},
}

@inproceedings{
generalagent,
title={General agents need world models},
author={Jonathan Richens and Tom Everitt and David Abel},
booktitle={ICML},
year={2025},
}

@article{probing,
  title={Probing classifiers: Promises, shortcomings, and advances},
  author={Belinkov, Yonatan},
  journal={Computational Linguistics},
  volume={48},
  number={1},
  pages={207--219},
  year={2022}
}

@inproceedings{cluster,
  title={Towards automatic concept-based explanations},
  author={Ghorbani, Amirata and Wexler, James and Zou, James Y and Kim, Been},
  booktitle={NeurIPS},
  volume={32},
  year={2019}
}

@inproceedings{sae,
  title={Sparse autoencoders find highly interpretable features in language models},
  author={Huben, Robert and Cunningham, Hoagy and Smith, Logan and Ewart, Aidan and Sharkey, Lee},
  booktitle={ICLR},
  pages={7827--7845},
  year={2024}
}

@inproceedings{hep-jepa,
  title={HEP-JEPA: A foundation model for collider physics using joint embedding predictive architecture},
  author={Bardhan, Jai and Agrawal, Radhikesh and Tilak, Abhiram and Neeraj, Cyrin and Mitra, Subhadip},
  journal={ICLR workshop},
  year={2025}
}

@article{
graphjepa,
title={Graph-level Representation Learning with Joint-Embedding Predictive Architectures},
author={Geri Skenderi and Hang Li and Jiliang Tang and Marco Cristani},
journal={TMLR},
issn={2835-8856},
year={2025},
}

@inproceedings{ad-l-jepa,
  title={Self-supervised representation learning with joint embedding predictive architecture for automotive lidar object detection},
  author={Zhu, Haoran and Dong, Zhenyuan and Topollai, Kristi and Sha, Beiyao and Choromanska, Anna Ewa},
  booktitle={AAAI},
  volume={40},
  number={16},
  pages={13925--13933},
  year={2026}
}

@article{jepa-dna,
  title={Jepa-dna: Grounding genomic foundation models through joint-embedding predictive architectures},
  author={Larey, Ariel and Dahan, Elay and Bleiweiss, Amit and Kellerman, Raizy and Leib, Guy and Nayshool, Omri and Ofer, Dan and Zinger, Tal and Dominissini, Dan and Rechavi, Gideon and others},
  journal={arXiv:2602.17162},
  year={2026}
}

@inproceedings{vla-jepa,
  title={Vla-jepa: Enhancing vision-language-action model with latent world model},
  author={Sun, Jingwen and Zhang, Wenyao and Qi, Zekun and Ren, Shaojie and Liu, Zezhi and Zhu, Hanxin and Sun, Guangzhong and Jin, Xin and Chen, Zhibo},
  booktitle={ECCV},
  pages={478--497},
  year={2026},
}

@article{mts-jepa,
  title={MTS-JEPA: Multi-resolution joint-embedding predictive architecture for time-series anomaly prediction},
  author={He, Yanan and Wen, Yunshi and Wang, Xin and Ma, Tengfei},
  journal={arXiv:2602.04643},
  year={2026}
}

@InProceedings{oocdm,
  title = 	 {Learning Causal Dynamics Models in Object-Oriented Environments},
  author =       {Yu, Zhongwei and Ruan, Jingqing and Xing, Dengpeng},
  booktitle = 	 {ICML},
  pages = 	 {57597--57638},
  year = 	 {2024},
  volume = 	 {235},
  
}

@InProceedings{tdb,
  title = 	 {Learning Cognitive Maps from Transformer Representations for Efficient Planning in Partially Observed Environments},
  author =       {Dedieu, Antoine and Lehrach, Wolfgang and Zhou, Guangyao and George, Dileep and Lazaro-Gredilla, Miguel},
  booktitle = 	 {ICML},
  pages = 	 {10287--10316},
  year = 	 {2024},
  volume = 	 {235},
  }

@inproceedings{
causal-structure,
title={A Causal World Model Underlying Next Token Prediction: Exploring {GPT} in a Controlled Environment},
author={Raanan Yehezkel Rohekar and Yaniv Gurwicz and Sungduk Yu and Estelle Aflalo and Vasudev Lal},
booktitle={ICML},
year={2025},
}

@inproceedings{
sae+circuit,
title={Transformers Use Causal World Models in Maze-Solving Tasks},
author={Alexander F Spies and William Edwards and Michael Ivanitskiy and Adrians Skapars and Tilman R{\"a}uker and Katsumi Inoue and Alessandra Russo and Murray Shanahan},
booktitle={ICLR Workshop on World Models: Understanding, Modelling and Scaling},
year={2025},
}

@inproceedings{droid,
  title={Droid: A large-scale in-the-wild robot manipulation dataset},
  author={Khazatsky, Alexander and Pertsch, Karl and Nair, Suraj and Balakrishna, Ashwin and Dasari, Sudeep and Karamcheti, Siddharth and Nasiriany, Soroush and Srirama, Mohan Kumar and Chen, Lawrence Yunliang and Ellis, Kirsty and others},
  booktitle={RSS},
  year={2024}
}

@inproceedings{3d-jepa,
  title={Locate 3d: Real-world object localization via self-supervised learning in 3d},
  author={McVay, Paul and Arnaud, Sergio and Martin, Ada and Majumdar, Arjun and Jatavallabhula, Krishna Murthy and Thomas, Phillip and Partsey, Ruslan and Dugas, Daniel and Gejji, Abha and Sax, Alexander and others},
  booktitle={ICML},
  year={2025}
}

@inproceedings{sr3d,
  title={Referit3d: Neural listeners for fine-grained 3d object identification in real-world scenes},
  author={Achlioptas, Panos and Abdelreheem, Ahmed and Xia, Fei and Elhoseiny, Mohamed and Guibas, Leonidas},
  booktitle={ECCV},
  pages={422--440},
  year={2020},
}

@inproceedings{scanrefer,
  title={Scanrefer: 3d object localization in rgb-d scans using natural language},
  author={Chen, Dave Zhenyu and Chang, Angel X and Nie{\ss}ner, Matthias},
  booktitle={ECCV},
  pages={202--221},
  year={2020},
}

@inproceedings{jhu-vpt,
  title={A vision foundation model for cataract surgery using joint-embedding predictive architecture},
  author={Shah, Nisarg A and Xia, Mingze and Vijay, Subhasri and Sikder, Shameema and Vedula, S Swaroop and Patel, Vishal M},
  booktitle={Medical Imaging with Deep Learning},
  year={2025}
}

@inproceedings{cataract-101,
  author    = {Klaus Schoeffmann and
               Mario Taschwer and
               Stephanie Sarny and
               Bernd M{\"{u}}nzer and
               Manfred J{\"{u}}rgen Primus and
               Doris Putzgruber},
  editor    = {Pablo C{\'{e}}sar and
               Michael Zink and
               Niall Murray},
  title     = {Cataract-101: video dataset of 101 cataract surgeries},
  booktitle = {Proceedings of the 9th {ACM} Multimedia Systems Conference},
  pages     = {421--425},
  
  year      = {2018},
  
}

@inproceedings{gue,
  title={DNABERT-2: Efficient foundation model and benchmark for multi-species genomes},
  author={Zhou, Zhihan and Ji, Yanrong and Li, Weijian and Dutta, Pratik and Davuluri, Ramana and Liu, Han},
  booktitle={ICLR},
  pages={41642--41665},
  year={2024}
}

@article{psg-jepa,
  title={Is Forward Prediction Enough? Physical State Grounding for JEPA World Models},
  author={Yan, Haodong and Zhu, Jiaguan and Jia, Mingyuan and Yin, Ruiqing and He, Junjie and Zhong, Zhide and Li, Junfeng and Lu, Jinxuan and Li, Hengtao and Zhang, Tianran and others},
  journal={arXiv:2608.06799},
  year={2026}
}

@article{3d-jepa-hu,
  title={3d-jepa: A joint embedding predictive architecture for 3d self-supervised representation learning},
  author={Hu, Naiwen and Cheng, Haozhe and Xie, Yifan and Li, Shiqi and Zhu, Jihua},
  journal={arXiv:2409.15803},
  year={2024}
}

@inproceedings{mmjepa-ecg,
  title={mmJEPA-ECG: Cross-Posture Robust Contactless Electrocardiogram Monitoring via Millimeter Wave Radar Sensing},
  author={Liu, Ziyang and He, Siyuan and Liang, Feng and Huang, Chang and Zhong, Shuxin and Wu, Kaishun},
  booktitle={AAAI},
  volume={40},
  number={45},
  pages={38962--38970},
  year={2026}
}

@article{us-jepa,
  title={US-JEPA: A Joint Embedding Predictive Architecture for Medical Ultrasound},
  author={Radhachandran, Ashwath and Ivezi{\'c}, Vedrana and Athreya, Shreeram and Anilkumar, Ronit and Arnold, Corey W and Speier, William},
  journal={arXiv:2602.19322},
  year={2026}
}

@article{cell-jepa,
  title={Cell-jepa: Latent representation learning for single-cell transcriptomics},
  author={ElSheikh, Ali and Wang, Rui-Xi and Wu, Weimin and Wen, Yibo and Dibaeinia, Payam and Zhang, Jennifer Yuntong and Hu, Jerry Yao-Chieh and Knudson, Mei and Babu, Sudarshan and Sun, Shao-Hua and others},
  journal={arXiv:2602.02093},
  year={2026}
}

@inproceedings{attribution,
  title={Axiomatic attribution for deep networks},
  author={Sundararajan, Mukund and Taly, Ankur and Yan, Qiqi},
  booktitle={ICML},
  pages={3319--3328},
  year={2017},
}

@article{counterfactual,
  title={Counterfactual explanations without opening the black box: Automated decisions and the GDPR},
  author={Wachter, Sandra and Mittelstadt, Brent and Russell, Chris},
  journal={Harvard Journal of Law \& Technology},
  volume={31},
  pages={841},
  year={2017},
}

@inproceedings{concept,
  title={Interpretability beyond feature attribution: Quantitative testing with concept activation vectors (tcav)},
  author={Kim, Been and Wattenberg, Martin and Gilmer, Justin and Cai, Carrie and Wexler, James and Viegas, Fernanda and others},
  booktitle={ICML},
  pages={2668--2677},
  year={2018},
}

@inproceedings{probe,
  title={Understanding intermediate layers using linear classifier probes},
  author={Alain, Guillaume and Bengio, Yoshua},
  booktitle={ICLR workshop},
  year={2017}
}

@article{circuit,
  title={Zoom in: An introduction to circuits},
  author={Olah, Chris and Cammarata, Nick and Schubert, Ludwig and Goh, Gabriel and Petrov, Michael and Carter, Shan},
  journal={Distill},
  volume={5},
  number={3},
  pages={e00024--001},
  year={2020}
}

@inproceedings{causal,
  title={Locating and editing factual associations in gpt},
  author={Meng, Kevin and Bau, David and Andonian, Alex and Belinkov, Yonatan},
  booktitle={NeurIPS},
  volume={35},
  pages={17359--17372},
  year={2022}
}

@article{steering,
  title={Steering language models with activation engineering},
  author={Turner, Alexander Matt and Thiergart, Lisa and Leech, Gavin and Udell, David and Vazquez, Juan J and Mini, Ulisse and MacDiarmid, Monte},
  journal={arXiv:2308.10248},
  year={2023}
}

@article{6g-jepa,
  title={JEPA for AI-Native 6G: Predictive Representations and Open Challenges},
  author={Hassan, Sheikh Salman and Meer, Irshad A and Saifaldawla, Almoatssimbillah and Tun, Yan Kyaw and Ozger, Mustafa and Alsenwi, Madyan and Van Huynh, Nguyen and Lee, Woong-Hee and Stefanovic, Cedomir and Sellathurai, Mathini and others},
  journal={arXiv:2607.09798},
  year={2026}
}
\bibliographystyle{icml2026}


\end{document}